\documentclass[letterpaper]{ieeeconf}
\usepackage[utf8]{inputenc}
\usepackage{amsmath}
\usepackage{amssymb}
\usepackage{makecell} 
\usepackage{multirow}

\usepackage{amsthm}

\usepackage[scr=boondoxo]{mathalfa} 

\usepackage{balance} 

\usepackage[pdftex]{graphicx}
\graphicspath{{./figures/}}
\DeclareGraphicsExtensions{.pdf,.jpeg,.jpg,.png}

\makeatletter
\let\MYcaption\@makecaption
\makeatother

\usepackage[font=footnotesize]{subcaption}

\makeatletter
\let\@makecaption\MYcaption
\makeatother

\usepackage[usenames,dvipsnames]{xcolor}

\usepackage{siunitx} 

\usepackage{cite}

\title{
3D Ensemble-Based Online Oceanic Flow Field Estimation\\ for Underwater Glider Path Planning
}

\author{Felix~H.~Kong$^1$,
        K.~Y.~Cadmus~To$^1$,
        Gary Brassington$^2$,
        Stuart Anstee$^3$,
        and Robert~Fitch$^1$
    \thanks{This research is supported by an Australian Government Research Training Program (RTP) Scholarship, Australia's Defence Science and Technology Group, Australian Bureau of Meteorology, and the University of Technology Sydney.}
    \thanks{$^1$Authors are with the University of Technology Sydney, NSW 2007, Australia {\tt\footnotesize Cadmus.To@student.uts.edu.au} and {\tt\footnotesize \{Felix.Kong,Robert.Fitch\}@uts.edu.au}}
    \thanks{$^2$ Author is with the Australian Bureau of Meteorology {\tt\footnotesize gary.brassington@bom.gov.au}}
    \thanks{$^3$ Author is with the Defence Science and Technology Group, Department of Defence, Australia {\tt\footnotesize stuart.anstee@dst.defence.gov.au}}
}

\renewcommand{\vec}[1]{\mathbf{#1}}
\newcommand{\mat}[1]{\boldsymbol{#1}}

\newcommand{\R}{\mathbb{R}}

\DeclareMathOperator*{\argmin}{arg\,min}

\begin{document}

\maketitle
\begin{abstract}
Estimating ocean flow fields in 3D is a critical step in enabling the reliable operation of underwater gliders and other small, low-powered autonomous marine vehicles. 
Existing methods produce depth-averaged 2D layers arranged at discrete vertical intervals, but this type of estimation can lead to severe navigation errors. 
Based on the observation that real-world ocean currents exhibit relatively low velocity vertical components, we propose an accurate 3D estimator that extends our previous work in estimating 2D flow fields as a linear combination of basis flows. 
The proposed algorithm uses data from ensemble forecasting to build a set of 3D basis flows, and then iteratively updates basis coefficients using point measurements of underwater currents. 
We report results from experiments using actual ensemble forecasts and synthetic measurements to compare the performance of our method to the direct 3D extension of the previous work. 
These results show that our method produces estimates with dramatically lower error metrics, with and without measurement noise.
\end{abstract}

\section{Introduction}
Oceanic flow field estimation is an important precursor to path planning for autonomous marine vehicles. 
Underwater gliders are particularly dependent on accurate estimates because of their comparatively limited thrust, requiring the aid of ocean currents to reach an intended destination.
Conversely, the advantage of underwater gliders is that they consume little power and can achieve months-long deployments relevant to applications in ocean science~\cite{Rudnick2016}, defence~\cite{Nott2015}, and other industries.

While 3D flow field estimation has long been a topic of study in oceanography, obtaining and digesting meteorological flow field estimates in a form useful for practical robotics applications is less well understood.
The goal of this paper is to develop a 3D estimator that builds on the idea of augmenting available forecasts with online measurement data.
\begin{figure}[t]
    \centering
    \begin{subfigure}[b]{\columnwidth}
        \centering
        \includegraphics[width=0.75\linewidth]{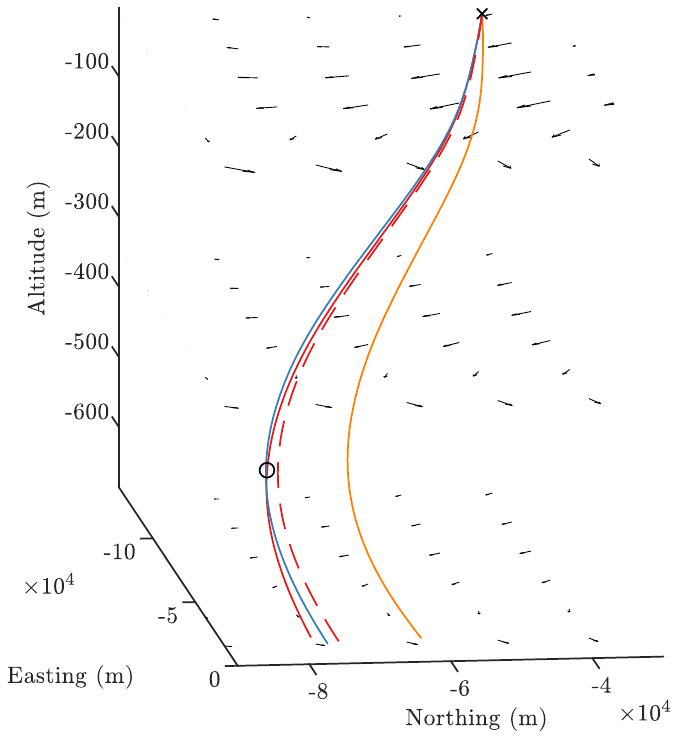}
        \caption{3D view}
        \label{fig:gliderHeadline_overview}
    \end{subfigure}
    \begin{subfigure}[b]{\columnwidth}
        \centering
        \includegraphics[width=\linewidth]{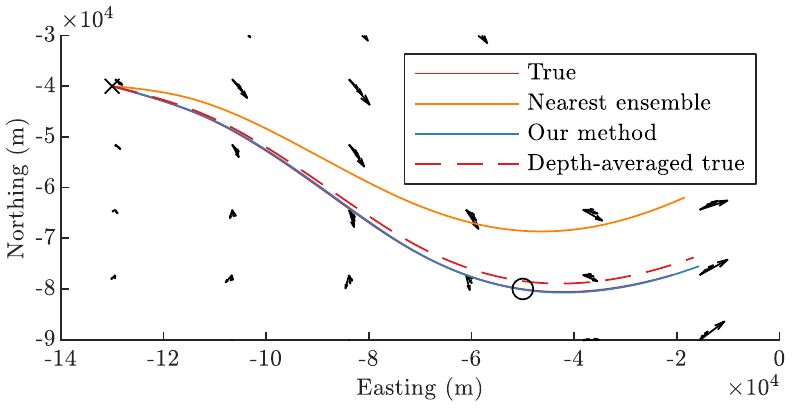}
        \caption{Top view}
        \label{subfig:gliderHeadline_top}
    \end{subfigure}
    \caption{Simulated glider trajectories when path-planning is performed using various flow field estimates. True flow field (black) is downsampled for ease of visualization. Departure location (at surface) marked by an `x', target destination (at 500m depth) marked by circle.}
    \label{fig:gliderHeadline}
    \vspace{-0.5em}
\end{figure}

The standard tool in meteorological forecasting is the \textit{ensemble forecast}, a collection of multiple predictions of future oceanic flow fields generated by simulation of ocean models.
Our previous work showed how data from an ensemble forecasting can be used to build a continuous flow field estimate in 2D \cite{Cadmus2021}.
However, for the purpose of path-planning for underwater gliders, which oscillate through a wide depth range, 2D estimates appear to be insufficient. 
As an illustration, Figure \ref{fig:gliderHeadline} shows glider trajectories planned based on various estimation approaches, including the common ``depth-averaged'' approach. 
Even when using the depth-averaged \textit{true} flow field for path planning, the error to the desired destination is visibly worse than the planning over our 3D-aware estimate generated by noisy measurement. 
This is due to the inability of the depth-averaged representation to account for depth-varying variations in the flow field. 
Hence accurate 3D flow field estimates are crucial to reliable performance of underwater gliders.

Our existing algorithm \cite{Cadmus2021} attempts to estimate an unknown 2D flow field as a linear combination of 2D ``basis flow fields'' generated from the ensemble forecast.
The direct extension to the 3D case would result in 3D basis flow fields, which are limited to the span of the ensemble forecast.
Crucially, the true flow field is unlikely to lie entirely within the span of the ensemble, which is a source of estimation error. 
To address this, a useful insight we make use of in this paper comes from noticing that 3D ocean flow sufficiently far from the coastline has negligible vertical velocity \cite{Munk1966,Roughan2002,Liang2017}. 
This fact allows 3D flow fields to be seen as a collection 2D horizontal flow fields.
Roughly speaking, building the basis directly from existing 3D flow fields would enforce the existing correlations in depth.
By breaking them apart into 2D flow fields and sharing these 2D flow fields between depths, the number of degrees of freedom increase, and greater expressivity of the basis is achieved.
The contribution of this paper is to not only to simply extend \cite{Cadmus2021} to 3D, but to adjust the ``naive'' 3D extension using the fact of negligible vertical velocity in order to increase the span of the basis, and hence improve estimation accuracy.

We describe the resulting algorithm in detail in Sections \ref{sec:problem_formulation} and \ref{sec:approach}, and evaluate its performance using ensemble forecast data produced by the Australian Bureau of Meteorology for an area of the Tasman Sea between Australia and New Zealand. We report results from experiments that compare the output of our algorithm with the ``naive'' 3D extension of \cite{Cadmus2021}, with and without measurement noise, and illustrate the accuracy of our method in Section \ref{sec:results_and_discussion}. 
In several cases, the root-mean-square (RMS) error of the achieved error of the proposed algorithm fell below the RMS of the lower bound of the ``naive'' 3D extension.

The algorithms and results we contribute are a promising step forward in understanding how forecasts of underwater current can be used to produce useful flow field estimates for planning. Although this work focuses on the case of time-invariant flows, it lays a foundation to begin to address the challenges of the time-varying case.

\section{Related work}\label{sec:related_work}
As stated previously, this work is an extension of a previous paper \cite{Cadmus2021}, which is the closest related work. 
This paper builds upon \cite{Cadmus2021} by extending to the case of 3D flow fields, and more importantly, by altering the ``naive'' 3D extension of algorithm to take advantage of the layered structure of 3D flow fields, resulting in improved accuracy.

This paper also joins a large body of work in the algorithmic estimation of flow fields. 
A standard tool for the estimation of flow fields in meteorology is the ensemble Kalman filter (e.g. \cite{Sakov2020,Evensen2003,Houtekamer1999}).
Similar to this algorithm, the ensemble Kalman filter takes as input an ensemble forecast~\cite{Leutbecher2008,gneiting2005weather}, and makes use of the well-known Kalman filter for prediction. 
However, the proposed algorithm in this paper is lightweight enough to directly use the Kalman filter update equations, instead of requiring a still-expensive Monte Carlo approximation of the covariance matrix update. 

Another class of flow field estimation techniques rely on direct simulation of a dynamical model of the ocean (e.g. \cite{Oke2013,Madec2008}). 
Many such models result in accurate estimates, not just of ocean currents, but also salinity, sea surface temperature, and many other important physical quantities. 
In fact, these methods are used to produce the very ensembles the present work considers as an input.
However, they are especially computationally intensive, and may be unsuitable for robotics applications that require relatively fast, up-to-date estimates.

In contrast to the above model-based techniques, some data-driven techniques have also previously been used for flow field estimation.
The ``incompressible Gaussian Process (GP)'' was introduced in \cite{Brian2019}, where a novel kernel was developed to enforce incompressibility of 2D flow field estimates.
Another related method, the kernel observer, combines a kernel embedding with an observer such as a Kalman filter \cite{Kingravi2016,Mardia1998,Whitman2017}.
While the present work makes use of a kernel embedding and a Kalman filter, it additionally uses an ensemble forecast and a singular value decomposition~(SVD).
The use of the SVD allow intuitive interpretability; also, these methods are not directly compatible with the ensemble format of meteorological forecasts.

Another class of data-driven methods used in estimation of fluid flows is the Dynamic Mode Decomposition~(DMD)~\cite{Schmid2010,Kutz2016,Brunton2019}, and its extensions (e.g. \cite{Williams2015,Williams2016}). 
These methods are based on a clever use of the SVD to obtain an approximate linear system model for fluid flows. 
The use of observers has also been proposed in conjunction with DMD \cite{Nonomura2018,Tu2013}. However, in these methods, the SVD is computed across samples in time, while in our method, the SVD is computed across ensemble members. 
Additionally, DMD only estimates fluid flow at the grid points (i.e. data locations); our method uses a kernel embedding to ``interpolate'' between grid points in a physically consistent way.


\section{Problem formulation} \label{sec:problem_formulation}
We consider the problem of producing the ``best'' estimate of an unknown 3D time-invariant flow field.
The inputs to our problem are an ensemble forecast, and a sequence of point measurements of the unknown flow field.

Suppose the ensemble forecast $\mathcal E$ is composed of $E$ ``members''. 
Each ensemble member $\mathscr e_i$, with $i\in\{1,\ldots, E\}$ is a flow field on a grid $\mathcal G\subset\R^3$ with $X\times Y\times Z$ points, so that $\mathcal E = \{\mathscr e_i | i  = 1, \ldots, E\}$.
Within each $\mathscr e_i$, associated with each grid point $x_g\in\mathcal G$ is a 2D velocity vector $u_i(x_g)$ representing the ocean's current in the horizontal direction, and assume the vertical velocity is negligible. 
Indeed, measurements and best estimates place the vertical velocity of the ocean's current on the order of $10^{-5}$\,\si{m/s} or less~\cite{Liang2017,Roughan2002,Munk1966}.
In contrast, the horizontal current in our dataset is on the order $10^{-1}$ to $10^0$\,\si{m/s}.
Notice that the ensemble members $\mathscr e_i$ are \textit{not} measurements of the true flow field, but predictions based on simulations of ocean models. 

The data from the true flow field takes the form of point measurements of the ocean current. 
The true flow field is modeled as a continuous flow field~${f_{true}:\R^3 \to \R^2}$. 
Then, the point measurements take the form:
\begin{equation}
    z(x) = f_{true}(x) + \mathscr n,
\end{equation} where $x\in\R^3$ is a position in 3D space, $\mathscr n\in\R^2$ is a random variable representing sensor noise drawn from some sensor model, e.g. $\mathscr n\sim N(0,\Sigma_{\mathscr n}$) for some covariance $\Sigma_{\mathscr n}$.
Let the set of all measurements be written~$\mathcal Z$. 
Notice that $z(x)\in\R^2$ describes the \textit{horizontal} velocity at a 3D point $x$ in the ocean, since the vertical component of the velocity is negligible. 

Additionally, a common assumption when modeling ocean flow is that the flow field should be incompressible~\cite{Kowalik1993,Madec2008}, often referred to as the \textit{Boussinesq approximation}~(e.g.~\cite[\S7.7]{Stewart2008}). 
Hence, a desirable property of a flow field estimate~$\hat f(x)$ is $\nabla \cdot \hat f(x) = 0$, i.e. $\hat f(x)$ is incompressible.

Then, given a sequence of measurements $z(x_k)\in\mathcal Z$ at locations $x_k$, we would like to find ``the'' flow field function~$\hat f^\star(x):\R^3\to\R^2$ such that 
\begin{alignat}{2}
f^\star(x) = &\argmin_{\hat f(x)\in C^\infty}        &\qquad& \sum_{k = 1}^{|\mathcal Z|} \|z(x_k) - \hat f(x_k)\|^2_2,\label{eqn:residual}\\
&\mathrm{subject~to} &      & \nabla \cdot \hat f(x) = 0 \mathrm{~for~all~} x \in\R^2\nonumber
\end{alignat}
 where $\|\cdot\|_2$ is the $\mathcal L^2$ norm of a vector, $C^\infty(x)$ is the set of smooth 2-vector valued functions on $x$, and $|\mathcal Z|$ is the number of elements in the set $\mathcal Z$.
 
However, it is apparent that the optimisation problem~\eqref{eqn:residual} is severely underconstrained, as there are infinitely many smooth, nondivergent vector fields that can fit a finite number of measurements $\mathcal Z$.
To address this, our previous work \cite{Cadmus2021} imposed the form 
\begin{equation}\label{eqn:representer_theorem}
    \hat f(x) = \mat H(x) w,
\end{equation} where $\mat H(x)$ is a $2\times N$ matrix function and $w\in\R^{N}$.
That paper constructed the basis of incompressible flow fields~$\mat H(x)$ from ensemble data, so that $\hat f^\star(x)$ is constrained to the span of $\mat H(x)$. 


However, the choice of $\mat H(x)$ in \cite{Cadmus2021} restricts its span to be equal to the span of $\mathcal E$, which can result in ``out-of-span'' error since the true flow field is unlikely to lie in the span of $\mathcal E$.
Hence the problem in this paper is to choose $\mat H(x)$ to have an increased span compared to the method in \cite{Cadmus2021} in a physically meaningful way in order to reduce ``out-of-span'' error.

Judicious selection of the basis $\mat H(x)$ is critical; it is certainly possible to add random basis vectors to increase the span of $\mat H(x)$, but this would be inefficient. 
Our approach is to share flow field information across depths in the ensemble data as a source of likely useful basis flows.




\section{Approach} \label{sec:approach}
\begin{figure}[tb]
    \centering
    \includegraphics[width=\linewidth]{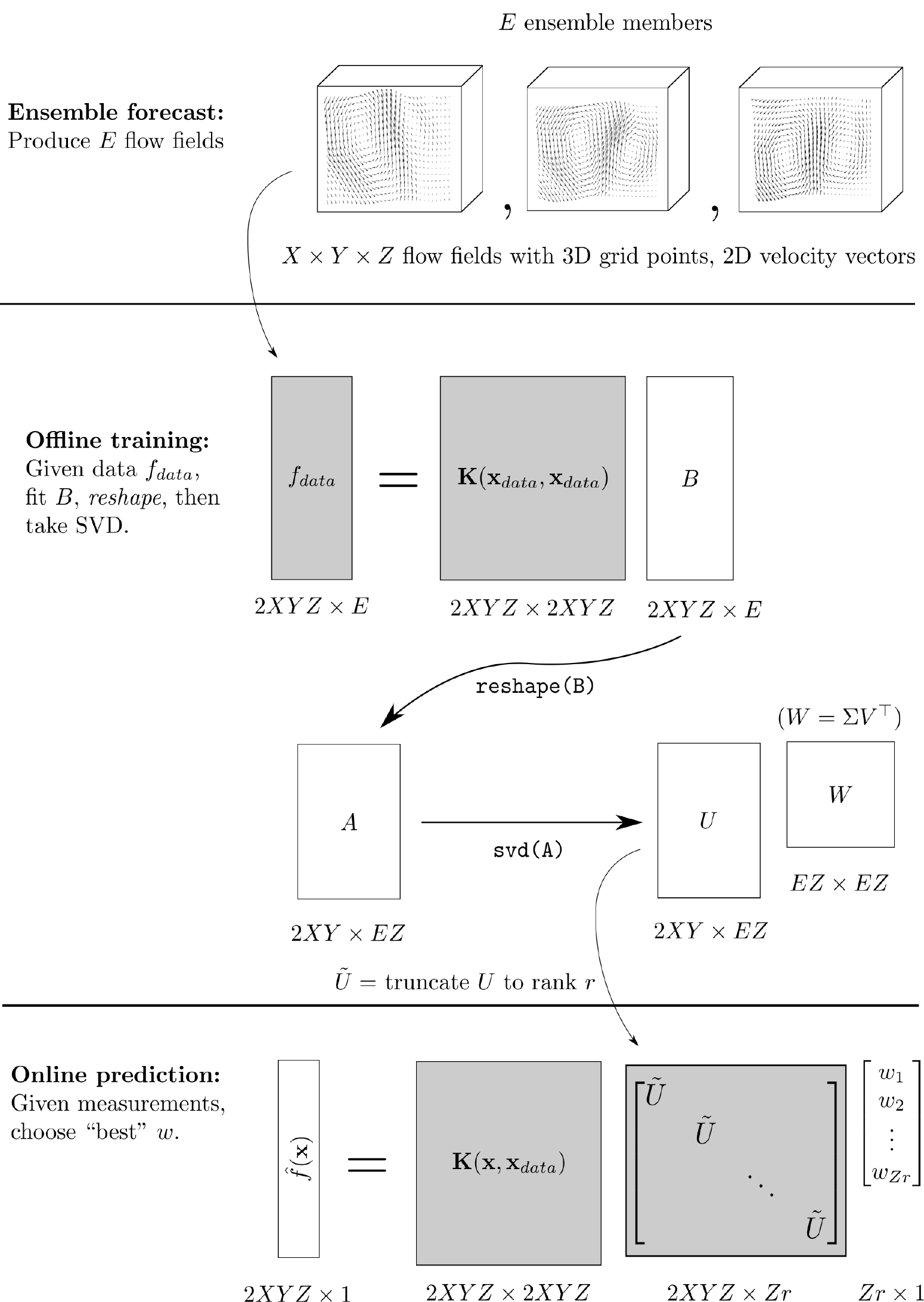}
    \caption{
        A sketch of the different components of $\hat f(x)$, showing the case of the thin SVD and where $S = ZE$.
        The key algorithmic contribution of this paper is the \texttt{reshape}, which allows many more modes (i.e. left singular vectors) in $\mat U$, resulting in a large span of the basis ${\mat H(\vec x) = \mat K(\vec x,\vec x_{data})\mat U}$.
        Shaded gray boxes indicate mathematical objects that are fixed at each stage.
    }
    \label{fig:algorithm_sketch}
    \vspace{-0.5em}
\end{figure}

In this section we describe the proposed algorithm, which is summarized in Figure \ref{fig:algorithm_sketch}.
As in \cite{Cadmus2021}, $\mat H(x)$ is computed offline based on ensemble data, and $w$ is updated recursively based on online measurements of the true flow field.


\begin{figure*}[tb]
    \centering
    \begin{minipage}{0.5\textwidth}
        \begin{subfigure}[b]{0.5\textwidth}
            \centering
            \includegraphics[width=\linewidth]{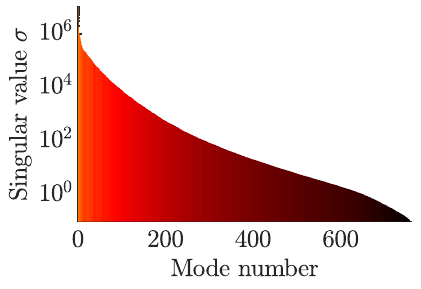}
            \caption{Singular values of $A$}
            \label{subfig:singular_values}
        \end{subfigure}\hfill
        \begin{subfigure}[b]{0.5\textwidth}
            \includegraphics[width=\linewidth]{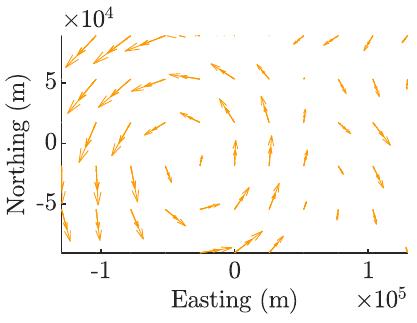}
            \caption{Top view of the first mode from $\mat H(\vec x)$}
            \label{subfig:basisExample_001_2D}
        \end{subfigure}
        \begin{subfigure}[b]{0.5\textwidth}
            \includegraphics[width=\linewidth]{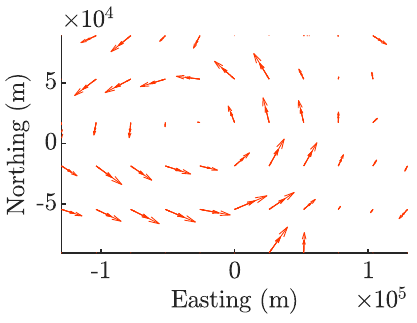}
            \caption{The 30th mode from $\mat H(\vec x)$}
            \label{subfig:basisExample_030_2D}
        \end{subfigure}\hfill
        \begin{subfigure}[b]{0.5\textwidth}
            \includegraphics[width=\linewidth]{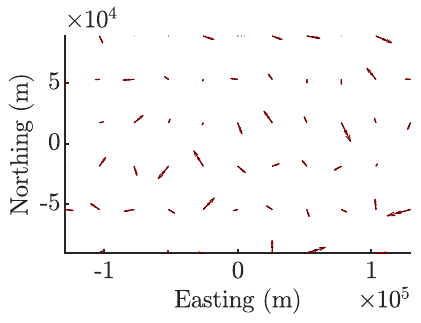}
            \caption{The 401st mode from $\mat H(\vec x)$}
            \label{subfig:basisExample_401_2D}
        \end{subfigure}
    \end{minipage}\hfill
    \begin{minipage}{0.5\textwidth}
        \begin{subfigure}[b]{\textwidth}
            \includegraphics[width=\linewidth]{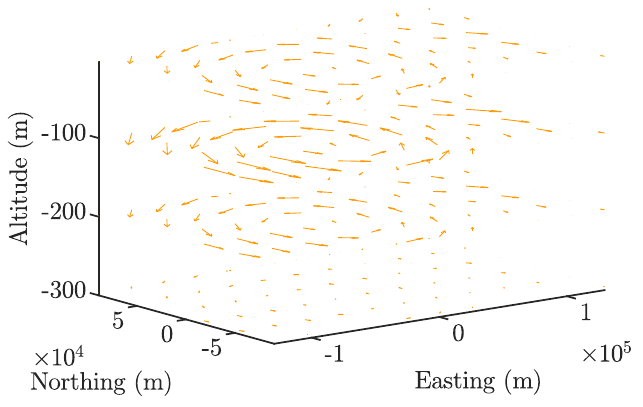}
            \caption{3D view of the first mode from $\mat H(\vec x)$}
            \label{subfig:basisExample_001_3D}
        \end{subfigure}
    \end{minipage}
    \caption{
        Example SVD of $A$.
        (\subref{subfig:singular_values}) shows the singular values corresponding to the modes of $\mat H(\vec x)$.
        Notice there are $S = 760$ modes, which is many more than $E = 95$ (the number that would have been produced by the ``naive 3D'' method, which is the direct 3D extension of \cite{Cadmus2021}).
        (\subref{subfig:basisExample_001_2D}-\subref{subfig:basisExample_001_3D}) shows some examples of the ``2.5D'' basis flow fields, corresponding to singular values of $1.0\times10^7$, $1.16\times10^5$, and $2.09\times10^1$, respectively. 
    }
    \label{fig:basis_flow_fields}
\end{figure*}

\subsection{Offline step 1: Kernel embedding}
Kernel methods are a common machine learning tool to allow linear methods to be applied to nonlinear patterns in data, and have previously been used to represent flow fields~\cite{Kingravi2016,Brian2019,Cadmus2021}. 
Kernels can be used to build a continuous spatial model of the flow field from data samples at discrete locations. 
The \textit{incompressible kernel} allows this ``interpolation'' to respect incompressibility \cite{Brian2019} in 2D; in this work we apply the incompressible kernel to describe 3D oceanic flow fields.

Given any kernel $k( x, x'):\R^3\times \R^3 \to \R$, the corresponding incompressible kernel is given by the following:
\begin{equation}
    K( x, x') = D( x)k( x,  x') D( x')^\top,
\end{equation} where $D( x) = [\frac{\partial}{\partial x_2}, - \frac{\partial}{\partial x_1}]^\top$, and $x_1, x_2$ are the first and second coordinate in the 3-dimensional vector $x$.
Notice that while ${x, x'\in\R^3}$, ${K(x, x')\in\R^{2\times 2}}$, as it describes the similarity between \textit{two-dimensional} flow vector components, and enforces 2D incompressibility. 
Due to the assumption of negligible vertical velocity, 2D incompressibility is sufficient for 3D incompressibility.

Let $\vec x_q$ be a vector of coordinates of $Q$ ``query locations''~$x_q$ in $\R^3$ with total length $3Q$, and $\vec x_{data}$ be the vector of all grid points $x_{data}\in\mathcal G$. 
Then, we write the kernel matrix 
\begin{equation}
    \mat K(\vec x_q, \vec x_{data}) = [K(x_q,x_{data})]\in\R^{2Q\times 2|\mathcal G|}
\end{equation} for all $x_q$ in $\vec x_q$ and $x_{data}$ in $\vec x_{data}$. The number of query points $Q$ is not required to match the number of points in $\mathcal G$.

Using this kernel matrix $\mat K$, inference can be performed.
For any $\beta\in\R^{2|\mathcal G|}$, the flow field 
\begin{equation}
    \hat f(x) = \mat K(x,\vec x_{data})\beta \label{eqn:f = Kbeta}
\end{equation} has the property that $\nabla \cdot \hat f(x) = 0$ for all $x\in\R^3$, i.e. it is incompressible.

Each ensemble member $\mathscr e_i$ can be represented by a latent representation $\beta_i$ by setting $\vec x_q = \vec x_{data}$ and performing the following linear regression:
\begin{equation}
    \beta_i = \argmin_\beta\|\vec u(\vec x_{data}) - \mat K(\vec x_{data}, \vec x_{data})\beta\|_2^2,
\end{equation}
where $\vec u_i(\vec x_{data})$ is the vector containing all the 2D velocity vectors $u(x_{data})$ for all $x_{data}\in\mathcal G$ in the $i$-th ensemble member $\mathscr e_i$.
Then, set of (finite) flow vectors $\mat K(\vec x_{data},\vec x_{data})\beta_i$ is the incompressible flow field that best fits $\mathscr e_i$ at the grid points in a least-squares sense.

In preparation for the next subsection, let $B$ be the $2|\mathcal G|\times E$ matrix 
\begin{equation}
    B = \begin{bmatrix}
    \beta_1 & \beta_2 & \cdots & \beta_E
    \end{bmatrix}.
\end{equation}

\subsection{Offline step 2: Singular value decomposition (SVD)}
The other major component of $\mat H(\vec x)$ comes from performing a singular value decomposition (SVD), which allow the columns of $\mat H(\vec x)$ to be easily interpretable as basis flow fields, which we call \textit{modes}.
The major contribution of this paper is to exploit the fact that 3D flow fields can be seen as a collection of depth-varying 2D flow fields, or equivalently, their 2D latent representations. 
Then, each 2D latent representation is used to generate a 3D flow field via $\mat K(x,\vec x_{data})$ that is self-similar in the depth dimension, which is added to the set of basis flow fields (Figure \ref{subfig:basisExample_001_3D} shows an example of one such basis flow field).
This results in up to $Z$ times the basis flow fields compared to using the 3D flow fields in the ensemble directly, thereby increasing the span, and hence the expressiveness of $\mat H(\vec x)$. 


To do this, we collect the horizontal slices of each $\beta_i$ in $B$ for the SVD for a shared ``library'' of basis vectors.
Each $\beta_i$~($2XYZ\times 1$) is a block of latent variables which can be reshaped into columns of horizontal slices
\begin{equation}
    \alpha_i = \begin{bmatrix}
    \beta_{i,z_0} & \beta_{i,z_1} & \cdots & \beta_{i,z_Z}
    \end{bmatrix}\in\R^{2XY\times Z},
\end{equation} where $\beta_{i,z}$ is the $2XY\times 1$ vector containing the elements of $\beta_i$ that correspond to a depth of $z$, and $\{z_0,z_1,\ldots,z_Z\}$ are the depth values present in the grid points $\mathcal G$.
Then, let the matrix $A$ be:
\begin{equation}
    A = \begin{bmatrix}
    \alpha_1 & \alpha_2 & \cdots & \alpha_E
    \end{bmatrix} \in \R^{2XY\times ZE}
\end{equation}

Throughout this paper we refer exclusively to the thin SVD, where only the entries corresponding to nonzero singular values are calculated.
The SVD of $A$ is:
\begin{equation}
    A = U\Sigma V^\top,
\end{equation} since $\mathscr e_i$, $K(x,x')$, and hence also $A$ are real-valued.
Notice that the number of columns in $U$ is $S = \min(ZE,2XY)$.
Typically, $E$ is limited, since producing ensemble forecasts is computationally expensive, but they are often computed over large grids, resulting in large $XY$ and $Z$. 
Hence in many cases, both $2XY$ and $ZE$ will be greater than $\min(E,2XYZ)$, the number of columns of $U$ resulting from the SVD of $B$.

Importantly, the SVD induces an ordering of the left singular vectors according to their singular values. 
In some cases, it can be advantageous to truncate $U$, $\Sigma$, and $V$ to reduce the data required to store $U$ by removing elements associated with small singular values.
Let $\tilde U$, $\tilde\Sigma$, and $\tilde V$ be the rank $r$ truncations of $U$, $\Sigma$, and $V$, respectively, with $r \leq S$.
Then, let $\mat U$ be:
\begin{equation}
    \mat U = \begin{bmatrix}
    \tilde U & & & \\
    & \tilde U & & \\
    & & \ddots & \\
    & & & \tilde U
    \end{bmatrix} \in \R^{2XYZ\times rZ},
\end{equation} with blank elements being equal to the appropriately sized zero matrices.

Finally, as in \cite{Cadmus2021}, the basis $\mat H(\vec x)$ is constructed in the following way:
\begin{equation}
    \mat H(x) = \mat K(x, \vec x_{data})\mat U. \label{eqn:H = KU}
\end{equation} 
Then, the weight vector $w\in\R^{rZ}$ selects a linear combination of the columns of $\mat H(x)$, which can be seen as 3D basis flow fields ``generated'' from a 2D flow field at a certain depth.
Figure \ref{subfig:basisExample_001_3D} shows a visualization of $\mat H(x,\vec x_{data})w_1$, where $w_1 = [1, 0, \ldots, 0]$.
The rest of Figure \ref{fig:basis_flow_fields} shows visualizations of several examples of the modes (i.e. columns) of $\mat H(x)$, and Figure \ref{subfig:singular_values} shows the relative contribution of each of these modes to the ensemble.

However, in cases where the number of modes is relatively high, significant truncation can be performed with surprisingly small loss in expressive power of $\mat H(x)$. Obviously, choosing $r < S$ obviously results in a reduction of the span of $U$, and hence $\mat H(x)$, so there is a tradeoff between compactness and expressibility.

Importantly, at each depth in $\mathcal G$, each $\tilde U$ will contain basis flow fields constructed using flow field information from \textit{all} depths.
 In addition, the 2D flow field at each depth has its own set  of weights, which may be distinct from other depths. 
This results in an increased span compared to the naive 3D extension of \cite{Cadmus2021}, where $\mat U$ is the left singular vectors of $B$, which would only have $E$ modes.

\subsection{Online update of $w$ via Kalman Filtering}
We now address the remaining part of \eqref{eqn:representer_theorem}, finding $w^\star$, which we determine based on online measurement data.
For the case of time-invariant flow fields considered in this paper, the Kalman Filter (KF) is not strictly necessary; however, it will be an invaluable tool when extending to the time-varying case. 

The well-known Kalman Filter recursively estimates the maximum likelihood state $w$ of a linear dynamical system, given a sequence of measurements $z_k\in\mathcal Z$ experiencing Gaussian sensor noise. 
In the case of time-invariant flow fields, the process model $F$ is simply the identity, and the predict step is:
\begin{align}
    w_{k+1} &= F w_k,\\
    \hat z_k &= \mat H(x_k)w_k + \mathscr n_k,
\end{align} where the subscript $k$ refers to the iteration number of the KF. 
The correction step equations are standard and omitted for brevity.

In the proposed algorithm, the initial state and covariance matrix of KF is initialised with the mean and variance of $W=\tilde{\Sigma}\tilde{V}^\top\in\R^{S\times ZE}$ across the ensemble members.
More verbosely, contiguous $Z$ columns of $W$ should be reshaped into new columns of $SZ$ values in the matrix $W^\prime\in\R^{SZ\times E}$.
The initial state $w_0$ is then computed as the the row-wise mean of $W^\prime$. 
The initial covariance matrix $P_0$ is set to be a diagonal matrix with diagonal elements equal to the row-wise variances of $W^\prime$.
Initialising the KF in this way expediently assumes that the estimated flow field is a random variable described by the same probability distribution from which the ensemble members are sampled.

\section{Simulation Results and discussion} \label{sec:results_and_discussion}
In this section, the proposed algorithm is applied to an example flow field estimation problem, and the ramifications for underwater glider path-planning are illustrated.

\subsection{Flow field estimation error comparison}
We first demonstrate the proposed algorithm on dataset provided by the Australian Bureau of Meteorology (BOM).
The ensemble forecast data provided by the BOM consists of 96 ensemble member flow fields forecasting conditions in the Tasman Sea between Australia and New Zealand on 16$^{\text{th}}$ of November 2018.
Notably, the forecast itself does not include data for the vertical velocity of the currents.
The flow field data covers a vast volume of ocean; we consider only a small region from $39.2^\circ$ to $40.8^\circ$\,S, $151^\circ$ to $154^\circ$\,E, and $2.5$\,\si{m} to $685$\,\si{m} deep.

The data is provided in geographic coordinates (latitude, longitude, and depth); an azimuthal projection about the center of the region was used to produce a Cartesian coordinate grid. 
To reduce computation time, this grid was then downsampled by a factor of 5 in the depth direction, resulting in a grid of size $\mathcal G$ is $31\times 17\times 8$. 

Since high-resolution sensor measurements of the region are unavailable, we extract one of the ensemble members and consider it to be the true unknown flow field. 
In the real world, the true flow field is unlikely to be one of the ensemble members, and may be significantly different.
In an attempt to capture this, the ``true'' ensemble member was hand-picked through visual inspection to be qualitatively different from the rest of the ensemble members.
The remaining 95 ensemble member flow fields form  $\mathcal E$.

\begin{figure}[tb]
    \centering
    \begin{subfigure}{\columnwidth}
        \includegraphics[width=\linewidth]{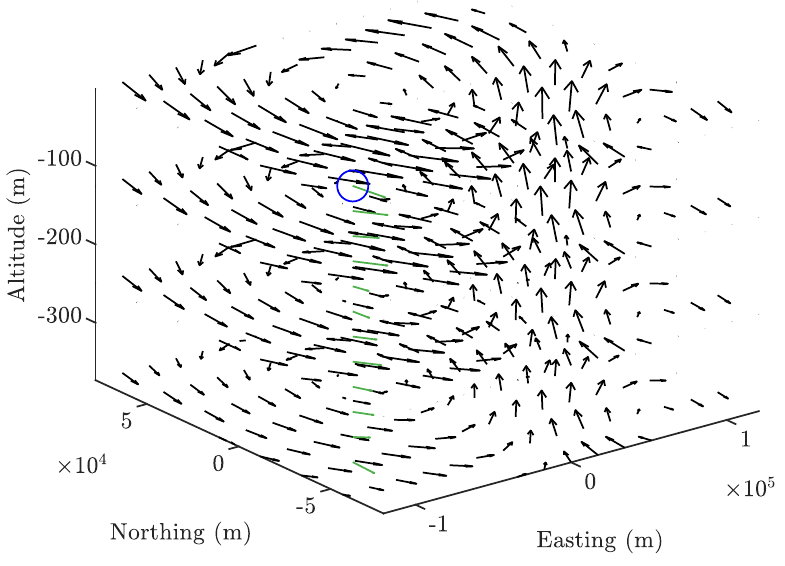}
        \caption{3D view of the environment with one ADCP ping}
    \end{subfigure}
    \begin{subfigure}{\columnwidth}
        \includegraphics[width=\linewidth]{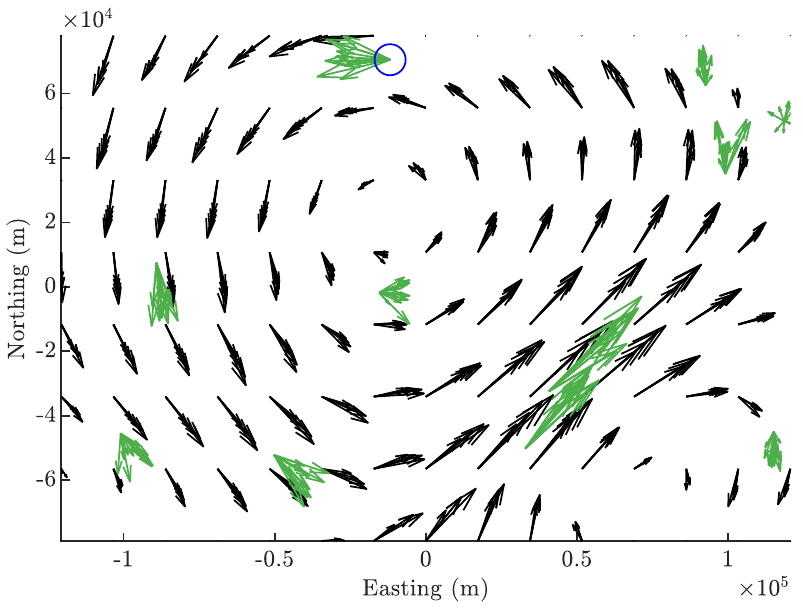}
        \caption{Top view of the environment with multiple ADCP pings}
        \label{subfig:more_measurement}
    \end{subfigure}
    \caption{
        Visualisation of measurements in the true flow field in black, downsampled by a factor of 2.
        A single ADCP ``ping'' results in 22 measurements (in green), only 12 of these are shown in (\subref{subfig:more_measurement}) for visibility.
        The location of a simulated ADCP-equipped surface vessel is shown in blue.
        Note the large noise on measurements.}
    \label{fig:true_flow_field_with_measurements}
\end{figure}

Simulated measurements were performed based on the true flow field $f_{true}(x)$, estimated via cubic interpolation. 
The way measurements were simulated was inspired by acoustic Doppler current profilers (ADCP), which are sensors that can measure ocean current in a column of water at regular depth intervals.
The simulated ADCP in our experiments has a vertical resolution of $32$\,\si{m} and a precision standard deviation of $9$\,\si{cm/s}.
It is worth noting that $9$\,\si{cm/s} standard deviation is quite large compared to the average speed of the current in the dataset, which was $22.8$\,\si{cm/s} (see the spread of the green arrows in Figure \ref{fig:true_flow_field_with_measurements}).

A radial basis function (RBF) is chosen as the kernel to reflect the property that in viscous fluid flow, motion on one ``layer'' of the fluid will exert an influence on nearby ``layers'':
\begin{equation}
    \label{eqn:rbf}
    k(x,x') = \sigma_k^2\exp(\langle x-x',[\ell_x^{-2}, \ell_y^{-2},\ell_z^{-2}]\rangle).
\end{equation} The hyperparameters were tuned by hand, with ${\ell_x = \ell_y = 10^4}$\,\si{m}, ${\ell_z = 100}$\,\si{m}, and $\sigma_k = \mu/\ell_x = 1718.9$\,\si{m^2/s}. 
The symbol $\mu$ is the average flow vector magnitude at the surface, where the flow is usually fastest.

Three methods were compared in simulated experiments: the proposed method, the ``naive 3D'' method, and the ``nearest'' ensemble member. Briefly, the naive 3D method is the direct extension of \cite{Cadmus2021} to the 3D case, performing the SVD on $B$ instead of on $A$, resulting in $E = 95$ modes, each representing a 3D flow field. The nearest ensemble member was the member whose estimated current at the measurement locations was the closest to the noisy measurements in a least-squares sense. 

Two sets of simulation experiments were performed.
To show the potential performance of the algorithm, the first set experienced `ideal' conditions with no measurement noise and measurement locations exactly at grid points.
The Kalman filter noise covariance was set to $R = 0.01^2I$, since there is no noise.
The second set encountered Gaussian noise with standard deviation $9$\,\si{cm/s}, and measurements were taken from approximately 450 randomly uniformly distributed surface locations for a total of 10,000 measurements. 
The Kalman filter noise covariance was increased to $R = 0.12^2I$ to account for both the the sensor noise model and the additional noise incurred from cubic interpolation during simulation of measurements between grid points.

Table \ref{tab:main_result_no_noise} and
Table \ref{tab:main_result} show the results the first and second set of experiments, respectively. 
The error shown in the tables is the error between the estimated and true flow fields evaluate at the grid points, irrespective of where the measurements were taken.
The lower bound of the RMS error of each method was obtained as the RMS error of $w^\star$, where
\begin{equation}
    w^\star = \argmin_{w\in\R^{rZ}} \sum_{x_g\in\mathcal G} \| u_{true}(x_g) - \mat H(x_g)w\|_2^2,
\end{equation} where $u_{true}(x_g)$ is the 2D flow velocity vector of the true flow field at $x_g\in\mathcal G$, which would be unknown in realistic situations. 
The RMS error of $w^\star$ serves as a measure of the error due to the true flow field being out of the span of $\mat H(x)$.

\begin{table}[tb]
    \centering
    \caption{Error performance and size of different methods\\ in ideal conditions}
    \begin{tabular}{c|c|c|c|c|c}
          & \makecell{No.\\ modes} & \makecell{No. elem\\ in $\mat H(x)$} & \makecell{Rel error\\ (\%)} & \makecell{RMSE\\ (\si{cm/s})} & \makecell{RMSE of\\ bound\\ (\si{cm/s})}\\
         \hline
        3D basis  & $95$ (3D) & $8.01\times10^{5}$    &      $8.30$ & $2.33$ & $2.33$\\
        \hline
        \multirow{5}{*}{2.5D bases} &  $760$ (2D) & $8.01\times10^{5}$      &    $1.42$ & $0.39$ & $0.077$\\
        & $400$ (2D)& $4.21\times10^{5}$  &  $1.78$ & $0.50$ & $0.34$ \\
        & $300$ (2D)& $3.16\times10^{5}$  &  $2.46$ & $0.69$ & $0.57$ \\
        & $200$ (2D)& $2.11\times10^{5}$ & $4.18$  & $1.17$ & $1.14$ \\
        & $95$ (2D)& $1.00\times10^{5}$ & $11.00$ & $3.08$ & $3.07$ \\
        \hline
        \makecell{Nearest $\mathscr e_i$\\$(i =82)$} & --- & ---- & $45.6$ &$12.80$&---\\
    \end{tabular}
    
    \label{tab:main_result_no_noise}
\end{table}

\begin{table}[tb]
    \centering
    \caption{Error performance and size of different methods with noise and random measurement location}
    \begin{tabular}{c|c|c|c|c}
         & \makecell{No.\\ modes} & \makecell{Rel error\\ (\%)} & \makecell{RMSE\\ (\si{cm/s})} & \makecell{RMSE of\\ bound\\ (\si{cm/s})}\\
         \hline
        3D basis  & $95$ (3D)    &      $11.11$ & $3.11$ & $2.33$ \\
        \hline
        \multirow{5}{*}{2.5D bases} &  $760$ (2D)       &    $9.81$ & $2.75$ & $0.077$ \\
        & $400$ (2D) & $9.91$ & $2.78$ & $0.34$ \\
        & $300$ (2D) & $10.64$ & $2.98$ & $0.57$ \\
        & $200$ (2D) & $11.96$  & $3.35$ & $1.14$ \\
        & $95$ (2D) & $17.15$ & $4.81$ & $3.07$ \\
    \end{tabular}
    
    \label{tab:main_result}
\end{table}

Immediately obvious from Table \ref{tab:main_result_no_noise} is that the proposed method (see rows labeled ``2.5D bases'') have consistently lower error bounds than the ``naive 3D'' method.
This is evidence that the proposed method increases the span of $\mat H(x)$ beyond the span of $\mathcal E$ in a meaningful way, allowing reduced out-of-span error. 

Not only are the \textit{bounds} of the proposed method lower than the bound of the naive 3D method, but so is the \textit{actual achieved} RMS error when 200 or more modes were used.
This illustrates that improved estimation performance can be achieved due to the increased expressiveness of $\mat H(x)$ in the proposed method. 

Figure \ref{subfig:singular_values} clarifies the reason why so many modes can be discarded during truncation while still retaining good representational ability of $\mat H(x)$. 
Noticing the logarithmic scaling of the y-axis, it can be seen that the last few hundred modes have a much smaller contribution compared to the first few.
For example, Figure \ref{subfig:basisExample_401_2D} appears to be mostly noise, and its contribution in Figure \ref{subfig:singular_values} is approximately $10^{-4}$ the dominant mode; hence the 401st mode can be omitted without much loss in representational power.
However, there does come a point where too many ``useful'' modes are truncated; we see in both Table \ref{tab:main_result_no_noise} and \ref{tab:main_result} that the estimation performance deteriorates as the number of modes drops to 200 or below.
Still, in both Tables \ref{tab:main_result_no_noise} and \ref{tab:main_result}, a significant number of entries in $\mat H(x)$ can be discarded while suffering only a small loss in estimation performance (see rows with 760 and 400 modes).

\subsection{Effect on glider planning performance}
In this subsection we show additional simulations which highlight the importance of accurate flow field estimation in path planning applications.

A simulated underwater glider mission was conducted in a sub-region of the region considered in the previous subsection. 
Inspired by mesopelagic ocean science missions~(e.g.~\cite{Yoerger2018,AndruszkiewiczAllan2020}), we consider an example glider mission where it is released from the surface and attempts to pass through a point at a depth of $500$\,\si{m}.
A constant glider velocity relative to flow is chosen by minimising the distance of the resulting path to the target in each flow estimation method, constrained to a fixed glider speed of $0.3$\,\si{m/s}.
Under the influence of the true flow field and this velocity, the motion of the glider was simulated using Euler integration with fixed spatial step size.

\begin{figure}[tb]
    \centering
    \begin{subfigure}[b]{\columnwidth}
        \centering
        \includegraphics[width=\linewidth]{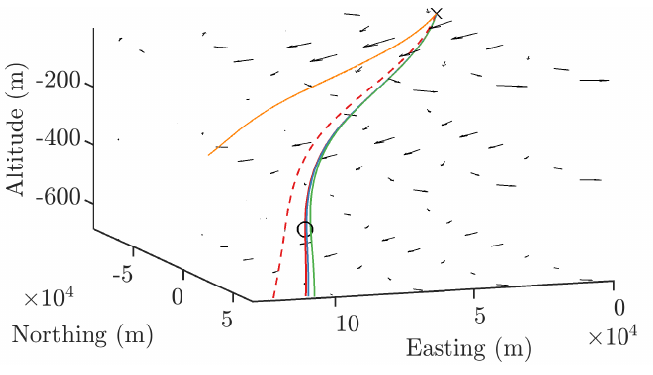}
        \caption{Overview}
        \label{subfig:gliderResults_overview}
    \end{subfigure}
    \begin{subfigure}[b]{\columnwidth}
        \centering
        \includegraphics[width=0.625\linewidth]{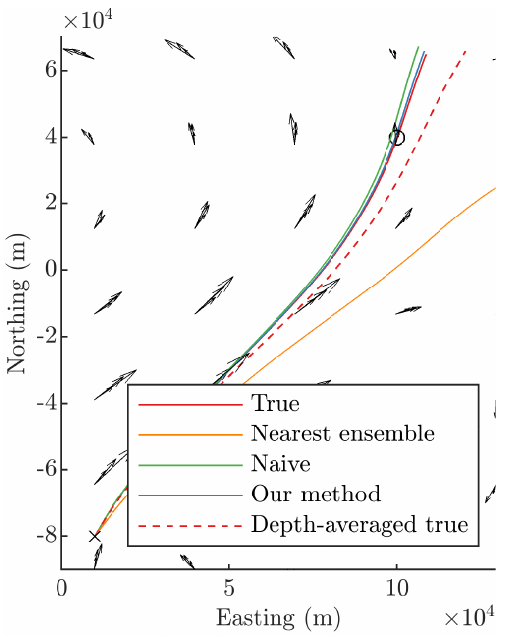}
        \caption{Top view}
        \label{subfig:gliderResults_top}
    \end{subfigure}
    \caption{Simulated glider trajectories with path-planning based on various flow field estimates. ``Our method'' refers the the proposed approach without truncation, and ``naive'' refers to the ``naive 3D'' case.}
    \label{fig:gliderResults}
\end{figure}

The resulting glider path for each flow field estimation method is shown in Figure \ref{fig:gliderResults}.
A glider trajectory generated by planning over the true flow field is given to show the ideal performance of the planner. 
``Our method'' refers to the untruncated 2.5D method generated with noise and unconstrained measurements, while the ``naive'' case is the naive 3D case under the same noise and measurement conditions.
It is clear that planning using the estimate generated by the proposed method is closest to ideal performance, followed by the naive 3D method.
We can see that a small difference of about 2\% in relative error between the 2.5D and 3D methods shown in Table \ref{tab:main_result}) results in a significant difference in error of $1303$\,\si{m}.

\begin{table}[tb]
    \centering
    \caption{Glider path results using controls planned with different methods}
    \begin{tabular}{c|c}
        & Closest approach (\si{m}) \\
        \hline
        True & $16$ \\
        Nearest ensemble & $29858$ \\
        Depth-averaged true & $6192$ \\
        Naive & $1808$ \\
        Our method & $505$ \\
    \end{tabular}
    \label{tab:glider_result}
    
\end{table}

Figure \ref{fig:gliderResults} also reinforces the idea that existing methods, i.e. depth-averaging, and choosing the nearest ensemble, are insufficient. 
For the purpose of glider path planning, depth-averaged flow fields are often used; its resulting trajectory suggests that even the depth-averaged \textit{true} flow field performs significantly worse than the 2.5D or 3D estimates. 
In practical situations, the true flow field will be unavailable; depth-averaging will further reduce the accuracy of any estimate it is performed on, negatively affecting glider performance.

Finally, another possible source of flow field data for glider path-planning is to simply use the nearest ensemble member~$\mathscr e_{82}$, which provides a 3D flow field. 
However, it performs the worst of all, suggesting that for the purpose glider path-planning, using the nearest ensemble is insufficient.

\section{Conclusion and Future Work}\label{sec:conclusion}
We have presented an algorithm for estimating 3D oceanic flow fields that exploits the property of negligible vertical velocity to produce highly accurate results. The most important avenue of future work is to further extend these ideas to address the time-varying case, which arises in long-duration deployments and planning long-distance paths. Another important direction for future work is to perform real-world validation of our algorithm using online observations.


\balance
\bibliographystyle{IEEEtran}
\bibliography{IEEEabrv,ref}

\end{document}